\documentclass{Interspeech}
\usepackage{amsmath,graphicx}
\usepackage{multirow}
\usepackage{xcolor}
\PassOptionsToPackage{colorlinks=true, linkcolor=green, citecolor=green, urlcolor=green}{hyperref}
\usepackage{hyperref}
\usepackage[noadjust]{cite}

\usepackage[normalem]{ulem}
\usepackage[final]{changes}

\interspeechcameraready

\title{SALM-Duplex: Efficient and Direct Duplex Modeling for Speech-to-Speech Language Model}

\author[affiliation={1}]{Ke}{Hu$^{*}$}
\author[affiliation={1}]{Ehsan}{Hosseini-Asl$^{*}$}
\author[affiliation={1}]{Chen}{Chen$^{*}$}
\author[affiliation={1}]{Edresson}{Casanova}
\author[affiliation={1}]{Subhankar}{Ghosh}
\author[affiliation={1}]{Piotr}{Żelasko}
\author[affiliation={1}]{Zhehuai}{Chen}
\author[affiliation={1}]{Jason}{Li}
\author[affiliation={1}]{Jagadeesh}{Balam}
\author[affiliation={1}]{Boris}{Ginsburg}

\affiliation{}{NVIDIA}{USA}
\email{kevinhu@nvidia.com}

\usepackage{comment}

\begin{document}

\maketitle

\begingroup
\renewcommand\thefootnote{*}
\footnotetext{Equal contribution.}
\endgroup

\begin{abstract}
    Spoken dialogue is an intuitive form of human-computer interaction, yet current speech language models often remain constrained to turn-based exchanges, lacking real-time adaptability such as user barge-in.
    We propose a novel duplex speech to speech (S2S) architecture featuring continuous user inputs and codec agent outputs with channel fusion that directly models simultaneous user and agent streams. 
    Using a pretrained streaming encoder for user input enables the first duplex S2S model without requiring speech pretrain. 
    Separate architectures for agent and user modeling facilitate codec fine-tuning for better agent voices and halve the bitrate (0.6 kbps) compared to previous works.
    Experimental results show that the proposed model outperforms previous duplex models in reasoning, turn-taking, and barge-in abilities. 
    The model requires significantly less speech data, as speech pretrain is skipped, which markedly simplifies the process of building a duplex S2S model from any LLMs. Finally, it is the first openly available duplex S2S model with training and inference code to foster reproducibility. Our code is open-sourced on GitHub\footnote{\scriptsize\url{https://github.com/cchen1436/NeMo/tree/main/examples/speechlm2}}.     
\end{abstract}

\vspace{-0.5em}
\section{Introduction}

Large language models (LLMs) \cite{brown2020language, team2024gemini, achiam2023gpt, dubey2024llama} have made significant strides in natural language processing, sparking interest in multimodal models that extend beyond text. Speech, as a natural interface for human-computer interaction, is a key part of this trend. Recent studies suggest adapting LLMs to process speech prompts for various speech-to-text (STT) tasks \cite{fathullah2024prompting, chu2023qwen, chen2024salm, kong2024audio, team2024gemini, dubey2024llama, hu2024chain}.

While traditional systems often respond with text, speech outputs are more intuitive for human-computer interaction. Cascaded spoken dialogue systems, like AudioGPT \cite{huang2024audiogpt}, use text as an intermediate representation, involving sequential modules such as ASR, LLM, and TTS. However, these systems face drawbacks like high latency, lack of interactive behaviors, and loss of paralinguistics. To address these issues, research has shifted towards end-to-end speech-to-speech (S2S) modeling.

Previous S2S models focus on half-duplex, turn-based interactions. For instance, SpeechGPT \cite{zhang2023speechgpt}, initialized from LLaMA, undergoes sequential fine-tuning on speech-only data and multimodal instruction sets to handle spoken question-answer (QA) tasks. Similarly, USDM \cite{kim2024unified} extends Mistral’s pretraining with interleaved speech-texet data for enhanced multimodal understanding. 
GLM-4-voice \cite{zeng2024glm} efficiently tokenizes speech using one codebook and large-scale speech-text pretraining for downstream tasks like ASR, TTS, and SQA.

Several pioneering or concurrent full-duplex S2S models have been recently proposed~\cite{defossez2024moshi, yusalmonn, chen2025minmo, zhang2024omniflatten}. However, these systems face increased complexity in model, data, and computation, which hinders their widespread research and adoption. The introduction of additional submodules for turn-taking between user and agent increases system complexity and reduces the end-to-end nature of the models. Moreover, the extensive speech-text pretraining required on top of the LLM backbone is resource-intensive and limits scalability to any LLMs. Finally, using codecs to model user and agent interactions simultaneously necessitates a delicate balance between speech perception and generation, presenting another significant challenge.

To tackle the above problems, we propose a novel duplex S2S system with the following contributions: 1) A novel duplex S2S architecture featuring continuous user inputs and codec agent outputs with channel fusion that directly models simultaneous text and speech of both the user and agent. 2) We demonstrate several key advantages over existing duplex models: The use of a pretrained encoder as input enables the first duplex S2S model \textbf{without speech pretraining} requirement; As the agent and user are modeled by the codec and the pretrained encoder separately,  this facilitates \textbf{codec fine-tuning} toward better agent voices. 3) We propose a set of systematic metrics to evaluate conversational behaviors such as turn-taking and barge-in. Finally, it is the first \textbf{open duplex S2S model} with both training and inference code publicly available to foster reproducibility.

\begin{figure*}[t]
    \centering
    \includegraphics[width=0.85\textwidth]{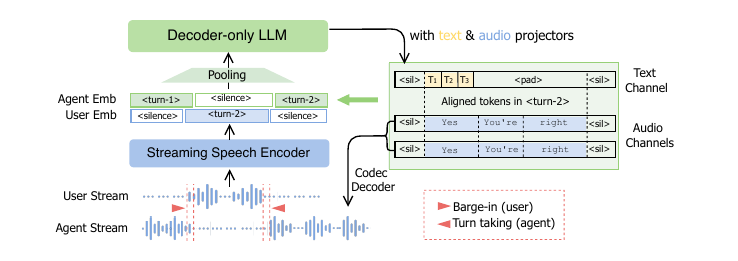}
     \vspace{-0.2em}
    \caption{The proposed duplex S2S model \textbf{without} requiring speech-text pretraining. Our model includes a streaming speech encoder, a personalized  codec, and an LLM. The model is trained to predict both text and audio channels in parallel with turn-level alignments.}
     \vspace{-1em}
    \label{fig:duplex_s2s}
\end{figure*}
\vspace{-1.2em}
\section{Related Work}

Interest in full-duplex S2S models has grown in the past year. Key challenges here include handling simultaneous user and agent streams and enabling turn-taking. Systems like \cite{wang2024freeze, xie2024mini, chen2025minmo} model single-channel interactions but use external signals, such as stopping commands \cite{xie2024mini} or submodules \cite{chen2025minmo}, to decide when to respond. Models like SyncLLM \cite{veluri2024beyond} and OmniFlatten \cite{zhang2024omniflatten} achieve full-duplex conversation by employing time chunking methods, embedding time information into LLMs for synchronization. This interleaving processing allows the model to handle user inputs like barge-in with low latency.


Our duplex S2S model is trained without speech-text  pretraining, unlike \cite{defossez2024moshi}. In multi-turn conversation, we align text and speech at the turn level, which simplifies data preparation compared to word-level alignment. Compared to \cite{yusalmonn, wang2024freeze, xie2024mini}, our model predicts text and speech simultaneously without requiring an explicit TTS component. Our speech codec model uses parallel codebooks (see details in Sec.~\ref{sec:codec}) and enables speech generation with minimal latency. Our design further enables codec fine-tuning for improved agent voices while halving the required bitrate of previous works (0.6 kbps).

\section{Model Architecture}


To achieve duplex behavior, our S2S model takes two input streams simultaneously: user speech stream, and agent speech and text stream. 
As shown in Fig. \ref{fig:duplex_s2s}, the user speech is first encoded to generate continuous embeddings by the speech encoder using an 80-ms frame rate. 
We use a 100M streaming speech encoder from a CTC model \cite{nvidia2023stt_fastconformer}. We initialize the backbone  LLM using the TinyLlama-1.1B-chat model \cite{zhang2024tinyllama}. A modality adapter is used between the speech encoder and the text LLM. 
To obtain the agent embeddings in training, we use a codec model \cite{casanova2025nano} to generate 12.5 Hz speech codes for the agent speech. LLM vocabulary is extended to include extra tokens from speech codec with zero initialization. 
The two inputs are time-aligned and summed as the input to the text LLM (similar to \cite{ma2024language}).
Both our speech encoder and text LLM are causal and thus streaming. In training, we fine-tune both the speech encoder and the backbone  LLM. Text and speech loss are weighted differently in training (see Sec.~\ref{subsec:training_details}). Our model is trained by multi-channel next token prediction similar to \cite{brown2020language}.

\subsection{Simultaneous Agent Text and Speech Prediction}

As shown in Fig. \ref{fig:duplex_s2s}, we encode speech using 4 codebooks at a rate of 12.5 frames per second \cite{casanova2025nano}, and text targets are tokenized into a separate channel. We align the text and speech tokens at the turn level based on their start time. We prepend separate \texttt{<BOS>} tokens for text and speech at the beginning of the turn and append \texttt{<EOS>} at the end of the turn. The gap between text and speech tokens are padded by text pad ID. We  also tried word-level alignment between text and speech as in~\cite{defossez2024moshi} and did not find improvement. Empirically, we find that the model tends to learn agent text first. Therefore, we introduce a small delay (i.e., one token) to the speech channels to better condition on text channel context without introducing significant latency.

\subsection{Personalization-friendly Speech Tokenization}
\label{sec:codec}

We employ a partially causal neural audio codec to transform raw speech signals into streaming tokenized representations. Given an audio signal $\mathbf{a}$, the codec generates a two-dimensional acoustic matrix, $\mathbf{C}_{T \times N} = \textit{CodecModel}(\mathbf{a})$, where $T$ denotes the downsampled sequence length, and $N$ represents the number of codebooks per timestep. Each element in $\mathbf{C}_{T \times N}$ is an $m$-bit discrete code. We adopt the state-of-the-art NanoCodec \cite{casanova2025nano}, which achieves reasonable-quality audio compression at 0.6 kbps with a frame rate of 12.5 frames per second, employing $N=4$ independent codebooks. The codec leverages Finite Scalar Quantization (FSQ) \cite{mentzer2023finite}, ensuring independence among codebooks. This independence removes the need for additional models or delay mechanisms, allowing all $N$ codebooks to be predicted in parallel at each timestep, thereby enabling fully parallel modeling with low latency. 

Our duplex design allows us to personalize the pretrained codec for agent voices to further enhance audio quality. This is enabled by modeling the agent and user separately with the speech codec and a pretrained causal speech encoder. In the experimental section, we will evaluate the benefits of  speech and reasoning quality resulting from codec personalization.

\section{Duplex Data for Training}
\label{sec:train-data}
\begin{table}[t]
  \centering
  \setlength{\tabcolsep}{2pt}
  \renewcommand{\arraystretch}{0.9}
  \small
\caption{Synthetic training data with multi-turn and barge-in.}
  \begin{tabular}{c | c |cc c c }
    \toprule
    Task & Dataset  & \#Hours & Speech & Multi-turn & Barge-in \\
    \midrule
     & ASR-QA & 20k &Mix & Augment & $\times$\\
     Spoken & MS MARCO & 0.2k & TTS & Augment & $\times$\\ 
  QA   & Alpaca & 0.2k & TTS & Augment & $\times$ \\
     & Internal SFT & 3k & TTS & Real & \checkmark \\ \midrule
    Conv- & UltraChat & 3k & TTS & Augment &\checkmark \\
    ersation& Topic & 0.3k & TTS  & Augment & \checkmark\\
    \bottomrule
  \end{tabular}
  \label{tab:training_data}
  \vspace{-1em}
\end{table}

Table \ref{tab:training_data} summarizes our training data which  can be categorized into  spoken QA and multi-turn conversations. 

\subsection{Single-turn synthetic and real spoken QA}
\label{sec:single}

\begin{figure}[t]
    \centering
    \includegraphics[width=\columnwidth]{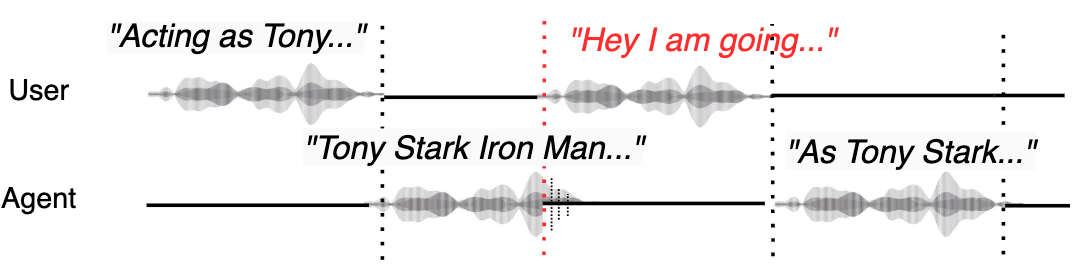}
    \caption{Duplex training data format. Our duplex data consists of separate user and agent streams including turn taking and barge-in behavior. Here, the user barges in at the second turn.}
    \label{fig:data}
    \vspace{-2em}
\end{figure}

Our most basic training data structure consists of a single-turn spoken QA between the user and agent. 
We use a multi-speaker TTS model~\cite{hussain2025koel} to synthesize the context, questions and answers from \textit{MS MARCO} \cite{bajaj2018human} and \textit{Alpaca} \cite{taori2023stanford}.
To mitigate overfitting to synthetic data, we follow~\cite{noroozi2024instruction} to create additional synthetic QA pairs using the Mixtral-8x22B LLM from English ASR-labeled data (8k public~\footnote{A subset from the NeMo ASR set in~\cite{nvidia2023stt_fastconformer}.} and 12k in-house). This data is then synthesized using the same TTS, denoted as \textit{ASR-QA}. The resultant user speech contains both TTS and real data. An evaluation set used in Sec.~\ref{sec:eval} is created from the public data portion. We use a fixed speaker to generate agent speech and randomly select speakers for user speech.



We create duplex training data from the aforementioned user-agent QA pairs. First, we split a pair of utterances into two streams, corresponding to the user and agent portions separately, and then insert silence into the agent stream when the user speaks, and vice versa. This gives us two streams of speech (shown as the first turn in Fig.\ref{fig:data}). This duplex structure enables the model to listen and speak simultaneously at any time. To prevent the agent from barging in unexpectedly, we insert a 0.64s silence between user and agent before the agent speaks. \added{This duration is chosen to balance modeling of user pauses with low turn-taking latency.}

\subsection{Augment with Multi-turn and Barge-in}

In order for the model to learn the ability for multi-turn conversation, we also create duplex data that includes two or more turns of conversation between the user and agent (e.g., Fig. \ref{fig:data}).
First, we synthesize 3k hours of duplex data from a text-based multi-turn \textit{Internal SFT} dataset to form multi-turn spoken QA.  
To ensure a more conversational flow, we limit each turn of the text SFT data, which is typically very long, to under 25 seconds.
Second, we augment the single-turn data from Sec.~\ref{sec:single} by randomly concatenating two QA pairs from the same dataset. 
The multi-turn data topics focus on role-playing, daily topics, scientific topics, etc.
Moreover, when creating multi-turn data, we  allow the user to barge in by cutting off the agent speech. After the cutoff, we keep a small duration (0.64 s) of agent speech to account for barge-in latency, and pad the rest of the agent turn with silence. As we show in later results, this straightforward approach enables the model to learn barge-in behavior.

\subsection{Conversational data}

To enhance the model's conversation ability on daily topics, we create \textit{Topic} and \textit{UltraChat} datasets (totaling 3.3k hours as shown in Table \ref{tab:training_data}). For both datasets, we first generate 4-turn text-based conversations and then synthesize them using a TTS model \cite{hussain2025koel}. For \textit{Topic}, we randomly choose a topic between user and agent and prompt the Meta-Llama-3.1-70B-Instruct model \cite{dubey2024llama} to generate a conversation. The topics are randomly chosen from the everyday-conversation dataset \cite{everydayconversations2024}, which covers 63 everyday and science topics. To generate concise replies for efficient training, we restrict the words of each turn to be 30 words in the prompt. The generated conversations are then synthesized into speech and prepared to the duplex data format. For \textit{UltraChat}, we randomly sample a chat conversation from the \textit{UltraChat} dataset \cite{ding2023enhancing} to use as contextual information in the prompt to generate a 4-turn conversation similar to \textit{Topic}. 

\section{Experiment Details}

\subsection{Training Details}
\label{subsec:training_details}

We implement the model with PyTorch using the NeMo Toolkit \cite{kuchaiev2019nemo}, and the model is trained on 32 A100 (80G) GPUs with a batch duration of 1000 sec per GPU. The speech encoder is initialized from a 100M streaming pretrained encoder with 80ms right context \cite{nvidia2023stt_fastconformer}, and the LLM is initialized from the 1.1B TinyLlama \cite{zhang2024tinyllama}. We use a 32k SentencePiece tokenizer for text, and a personalized 0.6 kbps NanoCodec \cite{casanova2025nano} for speech by default. Ablations for personalization are presented in Sec.~\ref{exp-codec}.
The speech codes have 4 channels, with a vocabulary size of 4,037 for each channel. 
Text and speech channel training loss are weighted by 3 and 1 respectively.
We use FusedAdam, and an inverse Square Root Annealing learning rate (LR) schedule for optimization. The LR schedule starts with an initial learning rate of 3e-4 with a warm-up of 2500 steps. Gradient clipping is applied at the threshold of 1.0 to stabilize training.

\begin{table*}[t]
\centering
\footnotesize
\vspace{-0.2cm}
\caption{Comparison of turn-taking and speech generation quality.}
\vspace{-0.2cm}
\resizebox{0.72\textwidth}{!}{%
\begin{tabular}{cc|ccc|c|c}
\toprule
 \multirow{2}{*}{Dataset}& \multirow{2}{*}{Model} & \multicolumn{3}{c|}{Barge-in Performance} & 1st  Response & UT \\ 
 & & Success $\ensuremath{\uparrow}$ & False Alarms $\ensuremath{\downarrow}$ & Latency (s) $\ensuremath{\downarrow}$ & Latency (s)  $\ensuremath{\downarrow}$ & MOS $\ensuremath{\uparrow}$ \\  \midrule
\multirow{2}{*}{UltraChat} & Ours & \textbf{83.0\%} & 0.0\% & \textbf{0.52} & 0.72 & \textbf{4.3} \\ 
 & Moshi & 56.0\% & 0.0\% & 0.63 & n/a & 3.9 \\ \midrule
 \multirow{2}{*}{Impatient} & Ours & \textbf{94.5\%} & 0.0\% & \textbf{0.69} & 0.92 & \textbf{4.0} \\ 
 & Moshi & 55.1\% & 0.0\% & 0.81 & n/a & 3.8 \\ \bottomrule
\end{tabular}}
\label{tab:conv}
\vspace{-0.5cm}
\end{table*}

\subsection{Evaluation Data and Metrics}
\label{sec:eval}
Our evaluation data consists of: 1) \textit{multi-turn} conversations: \textit{UltraChat}, \textit{Roleplay} (part of \textit{Internal SFT}), and \textit{Topic}, and 2) \textit{spoken QA reasoning}: \textit{ASR-QA} and \textit{Alpaca}. We select one shard for each dataset in Sec.~\ref{sec:train-data}, which is unseen during training, for this evaluation.
To evaluate model performance on a more challenging scenario where the user frequently interrupts the agent, we create an evaluation set called \textit{Impatient}. When creating \textit{Impatient}, we halve the silence time between the current and the next user turn (from the original duration in the \textit{ASR-QA} set) to increase the chance of the agent being interrupted by the user. By doing this, the interruption cases for our model and Moshi (more details in Sec. \ref{subsec:conv}) in the \textit{Impatient} dataset are as high as 95.4\% and 96.7\%, respectively.

In terms of evaluation metrics, we evaluate the reasoning ability of our model using GPT scores generated by {\tt\scriptsize gpt-4o-mini-2024-07-18} ranging from 0 to 10 based on the hypotheses and references of all the agent turns. The reasoning quality is evaluated using the aforementioned \textit{multi-turn} and \textit{spoken QA reasoning} datasets.
The hypotheses of agent turns are produced by transcribing the generated speech using the ASR model {\tt\scriptsize nvidia/parakeet-tdt\_ctc-110m}.

We evaluate turn-taking ability and speech generation quality using the \textit{UltraChat} and \textit{Impatient} datasets. We use two types of metrics to measure the turn-taking ability: barge-in performance and 1st response latency (see Table \ref{tab:conv}). For barge-in performance, we introduce the following metrics: 1) Barge-in latency: The time delay between the user's speech onset and the agent stopping its response; 2) Success rate: The percentage of cases where the agent successfully stops speaking within 1.5 seconds after the user interruption; and 3) False alarm rate: The frequency at which the agent incorrectly barges in while the user speaks. Additionally, if the user stops speaking within 0.1s, the event is not counted as a false alarm, as we found that Moshi tends to proactively respond. The 1st response latency is defined as the time taken by the agent to respond to the 1st user turn. To evaluate the speech quality, we compute the UTMOS \cite{saeki2022utmos}  using the generated agent speech  after removing silence.

\begin{table}[t]
\centering
\footnotesize
\caption{Reasoning quality of multi-turn conversation and spoken QA. GT+LLM denotes an optimal cascaded system which feeds every ground-truth user turn to the LLM.}
\vspace{-0.4cm}
\begin{tabular}{c|ccc|cc}
\toprule
\multirow{2}{*}{GPT Score} & \multicolumn{3}{c|}{Multiturn Conversation} & \multicolumn{2}{c}{Spoken QA} \\
 & UltraChat & Roleplay & Topic & ASR-QA & Alpaca \\ \midrule
Ours & \textbf{3.5} & \textbf{4.6} & \textbf{6.1} & \textbf{7.8} & \textbf{2.9} \\ 
Moshi & 3.4 & 1.7 & 2.8 & 1.9 & 1.7 \\ \midrule
GT+LLM & \dashuline{6.4} & \dashuline{4.9} & 5.5 & 5.8 & \dashuline{5.0} \\ \bottomrule            
\end{tabular}
\label{tab:reason}
\vspace{-0.8em}
\end{table}

\section{Results and Comparison}

\subsection{Conversation and Speech Generation Quality}
\label{subsec:conv}

We first evaluate the turn-taking and speech generation quality of our model in Table \ref{tab:conv}. 
Compared to Moshi, our model has significantly higher barge-in success rate (94.5\% v.s. 55.1\%), the same false alarm rates, and lower barge-in latency (0.69s v.s. 0.81s). 
We observe that, in multi-turn conversations, Moshi often initiates dialogue more proactively, leading to user barge-in failures for both \textit{UltraChat} and \textit{Impatient}. 

We cannot directly compare our  1st response latency with Moshi's as Moshi almost always responds before the user finishes talking and thus does not fit for this metric. We also note that our 1st response latency is affected by our data generation, as we always add a 0.64-second silence after the user turns to ensure no unexpected agent barge-in. \deleted{Further reducing this delay is our future work.}
Lastly, we report UTMOS and our model generates better quality speech than Moshi by up to 0.4.

\subsection{Reasoning Quality}

In Table \ref{tab:reason}, we compare the reasoning ability of our model to Moshi \cite{defossez2024moshi} and an {\em optimal cascaded system} formed by feeding every ground-truth user turn text to LLM (i.e., GT+LLM in Table \ref{tab:reason}). The backbone of our model, TinyLlama, is used as the LLM. We report the aforementioned GPT scores on two types of test sets: \textit{multi-turn conversation} and \textit{spoken QA}. Compared to Moshi, our model shows better scores on all datasets despite the fact that our model uses much less data and smaller backbone. 
Compared to the optimal cascaded system, our model shows competitive results, better on two and worse on three sets. The slightly worse performance of end-to-end versus cascaded is not new and has been shown by other research~\cite{defossez2024moshi,team2024gemini,zhang2023speechgpt,noroozi2024instruction}. Future works include i) a more fair comparison with full pipeline (VAD, streaming ASR and TTS, LLM), and ii) improving the reasoning of duplex S2S models.

\subsection{Speech Codec Personalization}
\label{exp-codec}

We personalize the codec to our agent voice by fine-tuning the codec on 21k ground-truth utterances from the target speaker. The model is evaluated on 228 test samples that are not seen during training. Perceptual quality is assessed using estimated Mean Opinion Scores (MOS) with Torchaudio-Squim \cite{kumar2023torchaudio}. Intelligibility is measured by computing the Character Error Rate (CER), comparing transcriptions from the Massively Multilingual Speech (MMS) model \cite{pratap2024scaling} for both ground-truth and reconstructed audio. Speaker similarity is evaluated using the Speaker Encoder Cosine Similarity (SECS) \cite{yourtts}, computed with the state-of-the-art ECAPA2 speaker encoder \cite{thienpondt2024ecapa2}.

\begin{table}[t]
\vspace{-0.2cm}
\caption{Evaluation of audio reconstruction and the resultant S2S quality across different codecs.}
\vspace{-0.2cm}
\label{tab:results-codec-ft}
\centering
\resizebox{0.48\textwidth}{!}{%
\begin{tabular}{lc|ccc|c}
\toprule
\multirow{2}{*}{Codec} &\scriptsize{Bitrate}&\multicolumn{3}{c|}{Audio Reconstruction}&\multicolumn{1}{c}{S2S} \\ 
  & \scriptsize{kbps} &  { MOS$\uparrow$} & {CER$\downarrow$} & {SECS$\uparrow$}  & \scriptsize{ASR-BLEU}$\uparrow$ \\ \midrule
 Mimi\cite{defossez2024moshi}   &   1.1      &   4.16        &   3.00   &  0.65  & n/a  \\ \midrule
Nano\cite{casanova2025nano}   &   1.2      &   4.67        &   1.44   &  0.77  & 18.1  \\
Nano\cite{casanova2025nano}   &   0.6      &    4.54       &   3.55   &  0.57  &16.2 \\
\ +personalized    &   0.6      &  \textbf{4.75}         & \textbf{1.36}      &  \textbf{0.94} &\textbf{18.7}  \\ \bottomrule
\end{tabular}
}
\vspace{-0.6cm}
\label{tab:results-tts}
\end{table}


Table \ref{tab:results-codec-ft} presents the evaluation results for the 1.1 kbps Mimi Codec \cite{defossez2024moshi}, 1.2 kbps, and 0.6 kbps versions of NanoCodec \cite{casanova2025nano}, and the proposed personalized version of 0.6 kbps NanoCodec. 
Personalization significantly enhances the performance of the 0.6 kbps NanoCodec. Notably, despite operating at nearly half the bitrate, our personalized codec outperforms both Mimi and NanoCodec at 1.2 kbps across all audio reconstruction metrics on the target speaker. 

As an ablation study, we further train our duplex S2S models with different codecs (last three rows in Table \ref{tab:results-codec-ft}). For simplicity, we report ASR-BLEU, which is calculated based on the reference agent texts and ASR transcripts of generated agent speech.
Results on \textit{ASR-QA} in Table \ref{tab:results-codec-ft} indicate that personalization enhances duplex modeling as well, leading to improved perceptual quality and higher BLEU scores.



\subsection{Listening Duplex Conversation Examples}

\begin{figure}[h!]
    \centering
    \includegraphics[width=\columnwidth]{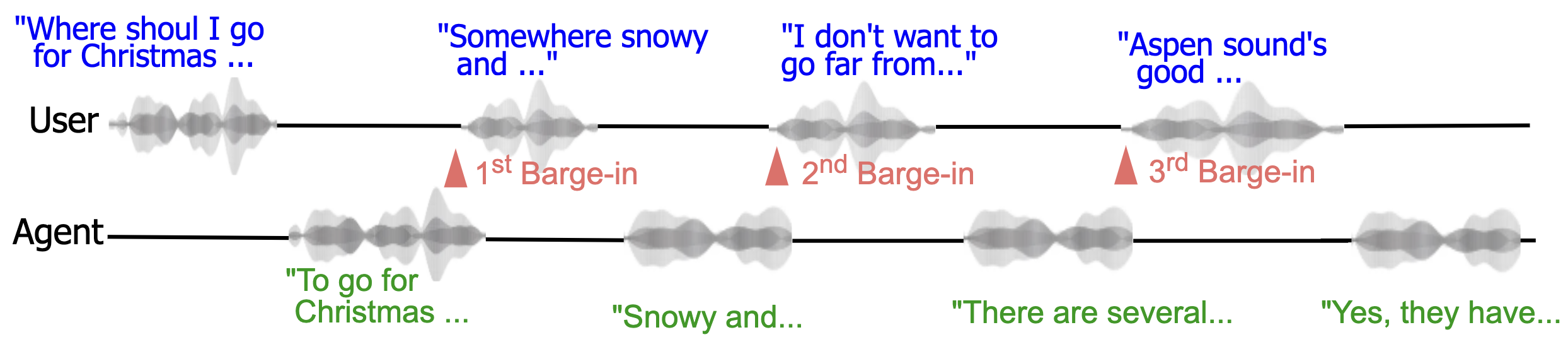}
    \caption{Multi-turn conversation with frequent barge-in.}
    \label{fig:demo1}
    \vspace{-1em}
\end{figure}

We include representative listening examples in an anonymous demo page\footnote{https://bit.ly/3FvVXcx}. 
Specifically, the following capabilities of our duplex S2S model on \textit{unseen} data are highlighted:

 \textbf{Robustness with frequent interruption}. In the example of Fig.~\ref{fig:demo1} and the webpage, the user interrupts the agent three times in 15 seconds, and leaves limited time for the agent to respond. Despite these challenges, the agent still demonstrates robust conversational behavior in handling frequent barge-in.

\begin{figure}[h!]
    \centering
    \includegraphics[width=0.75\columnwidth]{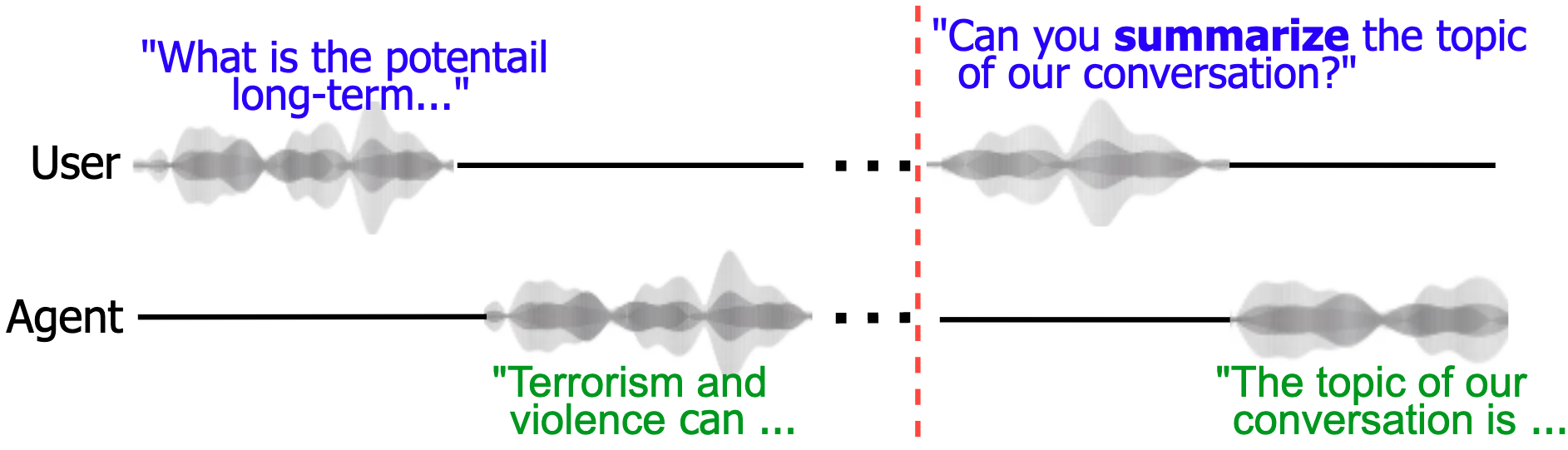}
    \caption{Spoken QA example on an unseen topic. }
    \label{fig:demo2}
    \vspace{-2em}
\end{figure}

\textbf{Unseen reasoning problem}.
 Beyond leveraging learned knowledge to generate responses, the agent also demonstrates the ability to utilize contextual information, effectively summarizing the main topic of each conversation in Fig.~\ref{fig:demo2} and webpage that was unseen during training. 

\section{Conclusion}

We introduced a novel duplex S2S architecture that models simultaneous user and agent streams without requiring speech pretraining. Our data-efficient approach maintains end-to-end modeling of conversation reasoning and behaviors. Experimental results show competitive performance in reasoning, barge-in, and turn-taking. Our open-sourced training and inference code will also be a valuable resource for future research.

\vfill\pagebreak

\bibliographystyle{IEEEtran}
\bibliography{ref}

\end{document}